\documentclass{article} % For LaTeX2e
\usepackage{so_preprint,times}

\usepackage{amsmath,amsfonts,bm}

\def\eqref#1{equation~\ref{#1}}
\def\1{\bm{1}}

\def\evb{{b}}

\DeclareMathAlphabet{\mathsfit}{\encodingdefault}{\sfdefault}{m}{sl}
\SetMathAlphabet{\mathsfit}{bold}{\encodingdefault}{\sfdefault}{bx}{n}

\def\gC{{\mathcal{C}}}

\def\sB{{\mathbb{B}}}

\usepackage{hyperref}
\usepackage{url}
\usepackage{algorithm}
\usepackage{algorithmic}
\usepackage{enumitem}
\usepackage{graphicx}
\usepackage{subfigure}
\usepackage{caption}

\usepackage{ulem}

\usepackage{ntheorem}
\theoremseparator{.}
\newtheorem{mydef}{Definition}
\newtheorem{mytheo}{Theorem}

\newenvironment{myproof}{{\noindent\it Proof.}\noindent}{\hfill $\square$\par}

\title{Structural Optimization Makes Graph Classification Simpler and Better}

\iclrfinalcopy

\author{Junran Wu, Jianhao Li, Yicheng Pan $^*$, Ke Xu \thanks{Correspondence to: Yicheng Pan$<$yichengp@buaa.edu.cn$>$, Ke Xu$<$kexu@buaa.edu.cn$>$} \\
State Key Lab of Software Development Environment \\
Beihang University \\
Beijing, 100191, China \\
\texttt{\{ryan\_wu, lijianhao, yichengp, kexu\}@buaa.edu.cn} \\
}

\newcommand{\vol}{\textrm{vol}}

\begin{document}
\maketitle

\begin{abstract}
In deep neural networks, better results can often be obtained by increasing the complexity of previously developed basic models.
However, it is unclear whether there is a way to boost performance by decreasing the complexity of such models.
Here, based on an optimization method, we investigate the feasibility of improving graph classification performance while simplifying the model learning process.
Inspired by progress in structural information assessment, we optimize the given data sample from graphs to encoding trees. In particular, we minimize the structural entropy of the transformed encoding tree to decode the key structure underlying a graph. This transformation is denoted as structural optimization.
Furthermore, we propose a novel feature combination scheme, termed hierarchical reporting, for encoding trees. In this scheme, features are transferred from leaf nodes to root nodes by following the hierarchical structures of encoding trees.
We then present an implementation of the scheme in a tree kernel and a convolutional network to perform graph classification.
The tree kernel follows label propagation in the Weisfeiler-Lehman (WL) subtree kernel, but it has a lower runtime complexity $O(n)$. The convolutional network is a special implementation of our tree kernel in the deep learning field and is called Encoding Tree Learning (ETL).
We empirically validate our tree kernel and convolutional network with several graph classification benchmarks and demonstrate that our methods achieve better performance and lower computational consumption than competing approaches.
\end{abstract}

\section{Introduction}
\label{sec:intro}
Over the years, deep learning has achieved great success in perception tasks, such as recognizing objects or understanding language, which are hard for traditional machine learning methods \citep{bengio2021deep}.
To further enhance performance, research efforts have generally been devoted to designing more complex models based on previously developed basic models; such improvements include increasing model depths (e.g., ResNet \citet{he2016deep}), integrating more complicated components (e.g., Transformer \citet{vaswani2017attention}) or even both (e.g., GPT3 \citet{brown2020language}). However, little work has focused on the research direction of boosting performance through simplifying the basic model learning process.

Similarly, there are many interesting tasks involving graphs that are hard to learn with normal deep learning models, which prefer data with a grid-like structure.
Graph neural networks (GNNs) have emerged and have recently been ubiquitous in terms of deep learning with graphs because of their ability to model structural information \citep{hamilton2017inductive, kipf2017semi, zhang2018end,xu2019powerful}.
In GNNs, each node recursively updates its feature vector through a message passing (or neighborhood aggregation) scheme, in which the feature vectors from neighbors are aggregated to compute a new node feature vector \citep{gilmer2017neural, xu2018representation}.
For tasks involving the overall characteristics, an entire graph representation can be obtained through a pooling operation \citep{xu2019powerful, ying2018hierarchical}.
To improve the performance of basic GNNs, various more complex models have been developed.
Considering the differences among graph nodes, attention mechanisms have been adopted in GNNs' message passing schemes to focus on more informative neighbors \citep{velivckovic2018graph, zhang2019adaptive}.
Regarding the pooling process, in addition to the basic sum or average pooling methods, more complicated pooling operations have been proposed for better learning with respect to entire graphs, such as SORTPOOL \citep{zhang2018end}, DIFFPOOL \citep{ying2018hierarchical}, STRUCTPOOL \citep{yuan2020structpool} and SOPOOL \citep{wang2020second}.
In this context, the improvement in performance comes with the price of model complexity, similar to the routine in deep learning. We argue that there is a way to boost performance while reducing the complexity of basic models.

Here, we investigate the feasibility of improving graph classification performance by simplifying model learning. Generally, given a problem, a simpler data structure comes with a simpler algorithm but is very likely to result in information loss and poor performance. Therefore, \emph{structural optimization}, which transforms the original structure of data into a simplified form while maintaining crucial features, is proposed. Different from the classic optimization, which optimizes the parameters of a given target (e.g., the parameters in a neural network are searched by minimizing the loss function), this optimization aims to search the simplified structure of data.
In addition, these retained features are expected to not only keep the key information of datasets but more importantly, many other features that negatively influence the given task can also be excluded. Consequently, this method could possibly yield a better result with even higher efficiency due to the simplified structure.

This work is inspired by structural entropy \citep{li2016structural,li2018mathematical},
a metric designed to assess the structural information of a graph.
Structural entropy can also be used to decode the key structure of a graph as a measure of the complexity of its hierarchical structure. In this paper, we realize structural optimization by transforming the given graph into a corresponding encoding tree that reflects the hierarchical organization of data. The crucial structural information underlying a graph can be kept in the encoding tree with minimal structural entropy.
Knowing that the essence of deep learning success is its superior feature characterization ability \citep{bengio2021deep}, our core view is that an encoding tree obtained after structural optimization is a much simpler data structure that also preserves the main features of the input graph.

Based on simplified encoding trees, we propose a novel feature combination scheme for graph classification, termed hierarchical reporting. In this scheme, we transfer features from leaf nodes to root nodes based on the hierarchical structures of the associated encoding trees.
We then present an implementation of the scheme in a tree kernel and a convolutional network, denoted as WL Encoding Tree (WL-ET) kernel and Encoding Tree Learning (ETL), to perform graph classification.
The tree kernel follows the label propagation in the WL subtree kernel but has a lower runtime complexity $O(n)$. ETL is a special implementation of our tree kernel in the deep learning field.
%such as the relationship between the Graph Isomorphism Network (GIN) and the WL subtree kernel.
Finally, we empirically validate our tree kernel and ETL on various graph classification datasets. Our tree kernel surpasses the state-of-the-art kernel-based methods and even outperforms GNNs on several benchmarks. Our ETL approach also achieves competitive performance relative to that of baselines.

We list our contributions in this work as follows:
\begin{itemize}
	\item We present a novel direction to boost the performance of learning models while reducing complexity with an optimization method.
	\item Based on structural optimization, we optimize the given data sample from graphs to encoding trees, which are much simpler data structures that have optimized feature characterization abilities.
	\item We develop a novel tree kernel (WL-ET) and a convolutional network (ETL) and empirically present their discriminative power on many graph classification benchmarks.
\end{itemize}

\section{Related work}
\label{sec:liter}
GNNs have achieved state-of-the-art results on various tasks with graphs, such as node classification \citep{velivckovic2018graph}, link prediction \citep{zhang2018link} and graph classification \citep{hamilton2017inductive, zhang2018end, xu2019powerful}. In this work, we devote our attention to graph classification scenarios.

% \subsection{Graph classification}
\textbf{Graph classification.} Graph classification involves identifying the characteristics of an entire graph and is universal in a variety of domains, such as urban computing \citep{bao2017planning}, social network analysis \citep{backstrom2011supervised}, chemoinformatics \citep{duvenaud2015convolutional}, bioinformatics \citep{borgwardt2005protein}, and even code analysis \citep{thost2020directed}.
In addition to previous techniques such as graph kernels \citep{shervashidze2011weisfeiler}, recently,
GNNs emerged and have become a popular way to handle graph tasks due to their effective and automatic extraction of graph structural information \citep{hamilton2017inductive, zhang2018end,xu2019powerful}.
To address the limitations of various GNN architectures, the GIN \citep{xu2019powerful} was presented in theoretical analyses regarding the expressive power of GNNs in terms of graph structure capture.
All GNNs are broadly based on a recursive message passing (or neighborhood aggregation) scheme, where each node recursively updates its feature vector with the ``message'' propagated from neighbors \citep{gilmer2017neural, xu2018representation}.
The feature vector representing an entire graph for graph classification can be obtained by a graph pooling scheme \citep{ying2018hierarchical}, such as the summation of all node feature vectors of the graph.
Accordingly, much effort has been devoted to exploiting graph pooling schemes, which are applied before the final classification step \citep{zhang2018end, ying2018hierarchical, yuan2020structpool, wang2020second}.
All these pooling methods help models achieve state-of-the-art results but increase the model complexity and the volume of computations.

% \subsection{Structural entropy}
\textbf{Structural entropy.} Structural entropy is a measure of the complexity of the hierarchical structure of a graph \citep{li2016structural,li2018mathematical}. 
% The encoding tree of a graph that achieves the minimum structural entropy indicates the optimal hierarchical structure. 
The structural entropy of a graph is defined as the average length of the codewords obtained under a specific encoding scheme for a random walk. That is, when a random walk takes one step from $u$ to $v$ and we use $v$'s codeword to label this step, the codeword of the longest common ancestor of $u$ and $v$ on the encoding tree, which is also their longest common prefix, is omitted. This shortens the average codeword length. Equivalently, the uncertainty of a random walk is characterized by this value, and it is why we call it structural entropy.
The encoding tree, which is considered to be the essential hierarchical structure of the given graph, is achieved when the structural entropy is minimized. Structural optimization is the task of searching for this optimum. For more information on structural entropy, please refer to \citep{li2016structural}. Two- and three-dimensional structural entropy, which measure the complexity of two- and three-level hierarchical structures, respectively, have been applied in bioinformatics \citep{li2018decoding}, medicine \citep{li2016three}, the structural robustness and security of networks \citep{li2016resistance}, etc.

Very recently, a novel hierarchical encoding algorithm based on structural entropy optimization was proposed \citep{pan2020information}. This algorithm stratifies a given graph into multiple levels by minimizing structural entropy, during which the sparsest levels of a graph are differentiated recursively. This provides an efficient approach to approximate the optimal hierarchical structure. In this paper, we apply this algorithm to structural optimization.

\section{Methodology}
\label{sec:sturct_opt}
Based on structural optimization, we analyze the feature characterization of a graph by optimizing its structure and carry out graph classification on the new simplified structure. Specifically, we analyze the hierarchical structure of a graph that an encoding tree incorporates. A tree is a much simpler data structure than its original graph, while a high-quality encoding tree retains the structural information of the graph. Next, we first present the structural optimization process that transforms a graph into its encoding tree for simplified feature characterization. Based on the optimized encoding trees, we propose a tree kernel and a corresponding implementation in a deep learning model for graph classification. We elaborate on them below.

\subsection{Structural optimization}
Given a weighted graph $G=(V,E,w)$ and an encoding tree $T$ for $G$, the \emph{structural entropy of $G$ on $T$} is defined as
\begin{equation}
\mathcal{H}^T (G)=-\sum_{\alpha\in T} \frac{g_\alpha}{\vol(V)} \log \frac{\vol(\alpha)}{\vol(\alpha^-)}.
%\footnote{For notational convenience, for the root $\lambda$ of $T$, we set $\lambda^-=\lambda$. Thus, the term for $\lambda$ in the summation is $0$.}
\end{equation}
We define the \emph{structural entropy of $G$} to be the minimum entropy among all encoding trees, and it is denoted by
$\mathcal{H}(G)=\min_{T}\{\mathcal{H}^T (G)\}.$
$\mathcal{H}^T (G)$ is essentially the optimal hierarchical structure in the sense that the average length of the codewords obtained with a random walk under the aforementioned encoding scheme is minimized.

To formulate a natural encoding tree with a certain height, we define the \emph{$k$-dimensional structural entropy of $G$} for any positive integer $k$ to be the minimum value among all encoding trees with heights of at most $k$:
\begin{equation}
\mathcal{H}^{(k)}(G)=\min_{T:\text{height}(T)\leq k}\{\mathcal{H}^T (G)\}.
\end{equation}
The algorithm proposed by \citep{pan2020information} is devoted to computing a $k$-dimensional encoding tree with minimum structural entropy. We use this algorithm for structural optimization, which yields an encoding tree $T$ with a height of at most $k$ for $G$, where $T=(V_T, E_T)$, $V_T=(V_T^0, \dots, V_T^k)$ and $V_T^0=V$. We denote $n=|V|$ and $m=|E|$. It is worth noting that although the time complexity of this algorithm is $O(m\log m+n^2)$ in the worst case, if we denote by $h_{\max}$ the maximum height among the binary trees that appear during the construction of $T$, the time complexity reduces to $O(m\log m+h_{\max} n)$. Since the minimized structural entropy tends to generate balanced encoding trees \citep{pan2020information}, $h_{\max}$ is usually of order $O(\log n)$, which reduces the time complexity of structural optimization to $O(m\log m)$ (almost linear in the number of edges). In this paper, we compare the real running times of structural optimization on all test datasets in Appendix \ref{sec:time_com} with other parts of our methods, and the results show that the time required for structural optimization is much less than that of WL-ET and ETL and only accounts for 0.002\% to 4\% of the ETL time requirement.

\subsection{Tree kernel for graph classification}
Following the construction of the WL subtree kernel \citep{shervashidze2011weisfeiler}, we propose a novel tree kernel that measures the similarity between encoding trees, named the WL-ET kernel. The key difference between the two kernels is the label propagation scheme, where we develop a \textit{hierarchical reporting} scheme to propagate labels from child nodes to their parents according to the hierarchical structures of encoding trees. Finally, our tree kernel also adopts the counts of the node labels at different heights of an encoding tree as the feature vector of the original graph.

\begin{figure}[!hb]
  \centering
  \subfigure[]{ 
    \includegraphics[width=.48\textwidth]{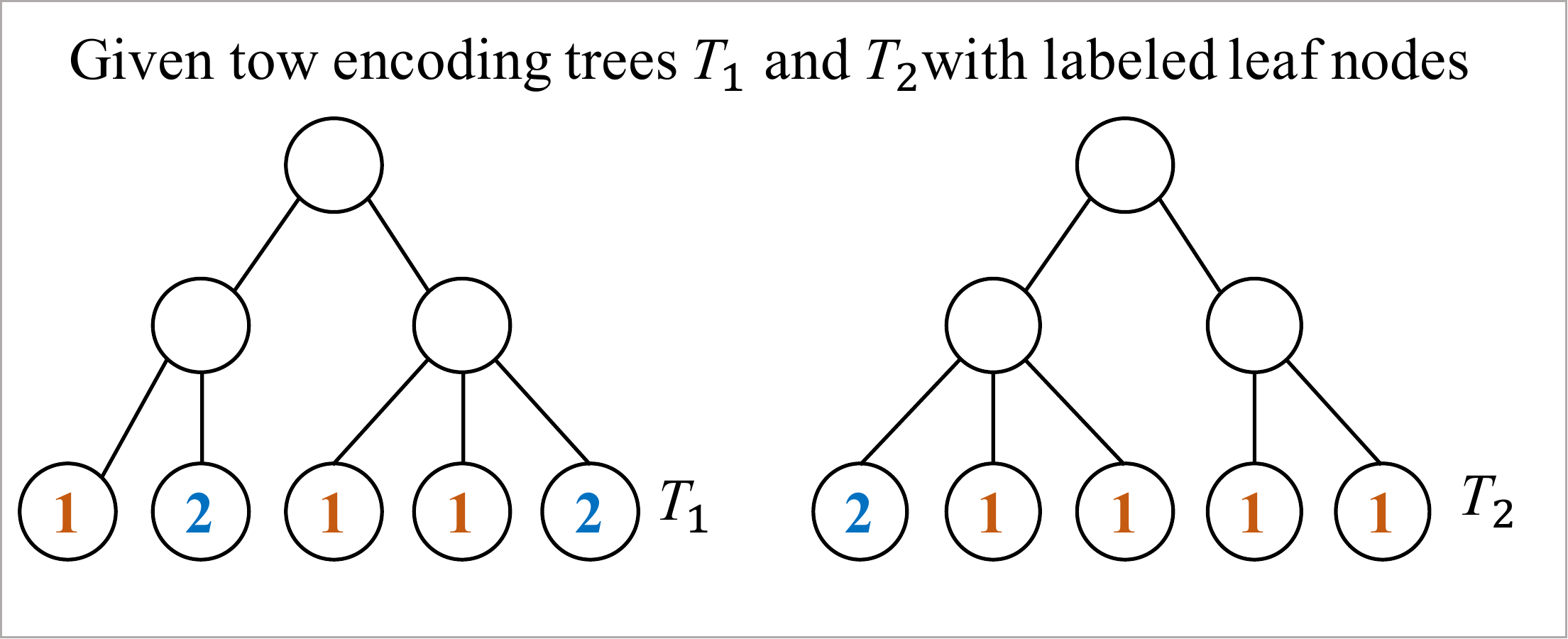} 
    \label{fig:treeKerenl_1} 
  }
  \subfigure[]{ 
    \includegraphics[width=.48\textwidth]{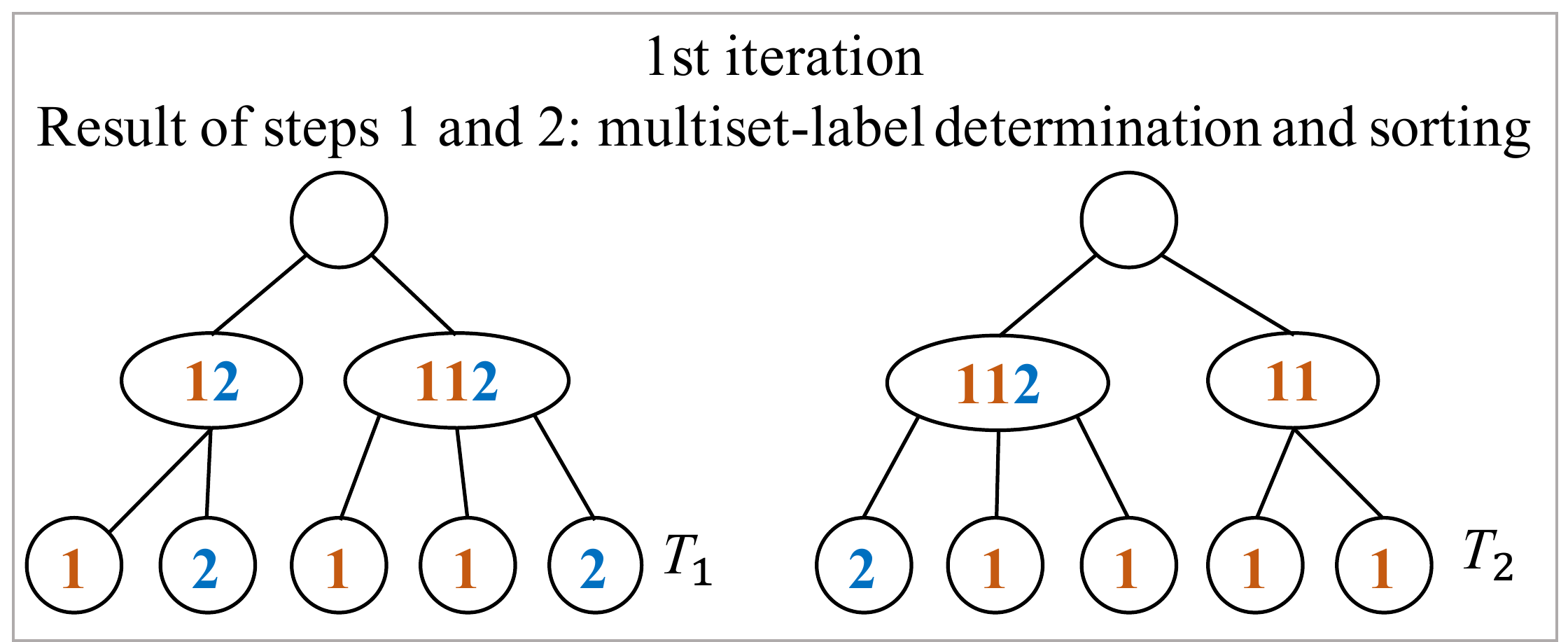} 
    \label{fig:treeKerenl_2} 
  } 
  \\
  \subfigure[]{ 
    \includegraphics[width=.48\textwidth]{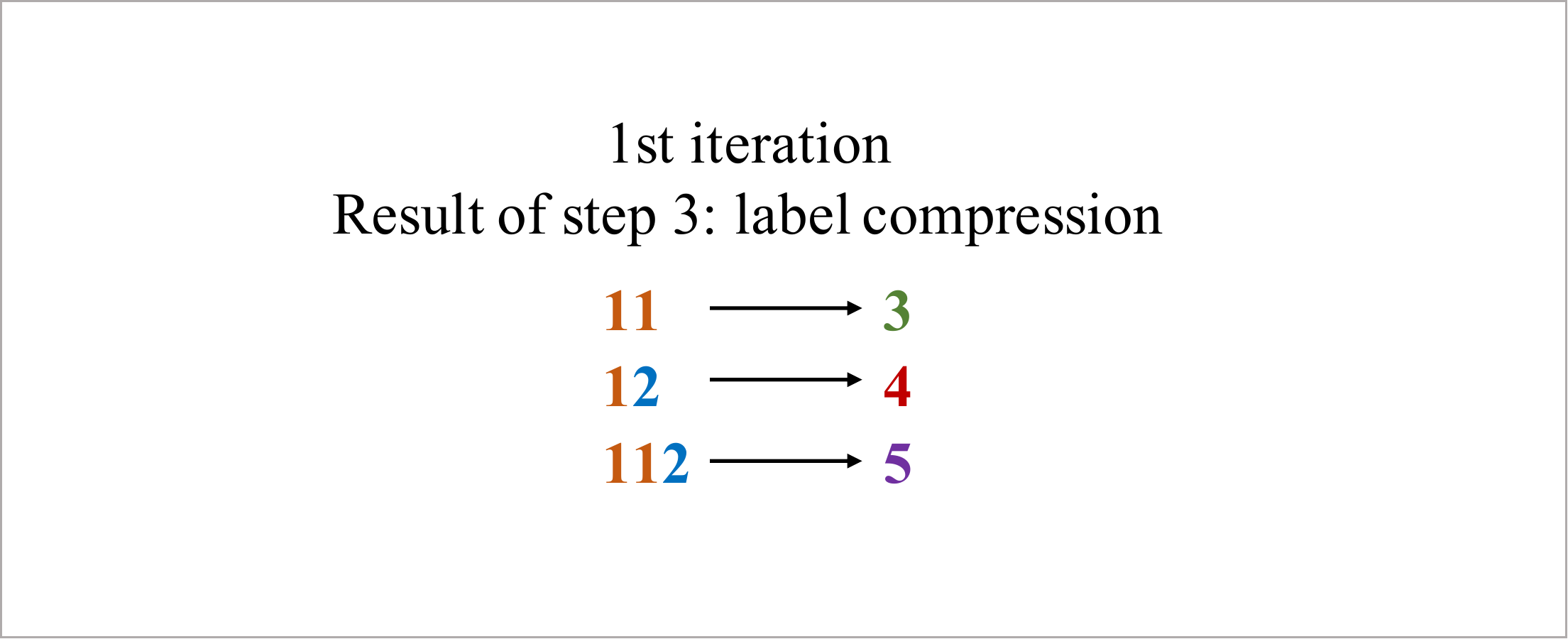} 
    \label{fig:treeKerenl_3} 
  } 
  \subfigure[]{ 
    \includegraphics[width=.48\textwidth]{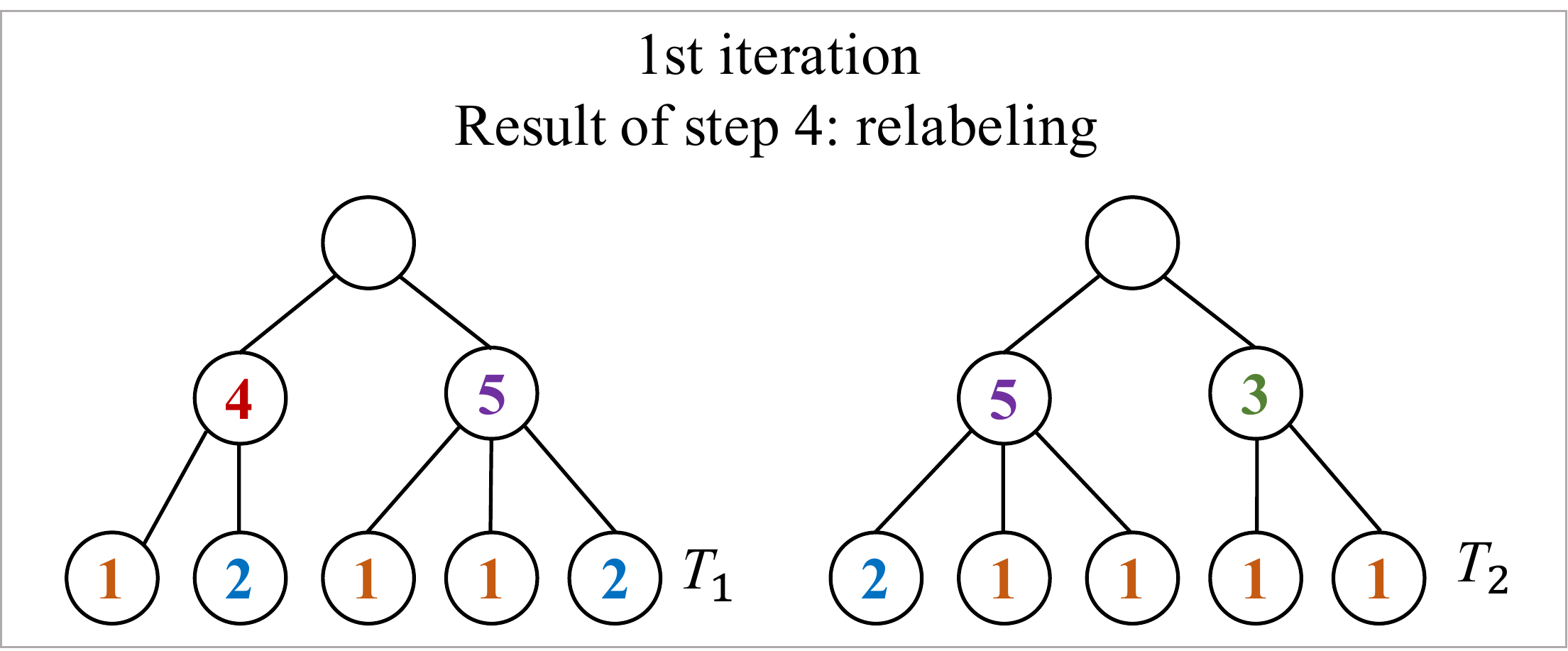} 
    \label{fig:treeKerenl_4} 
  } 
  \\ 
  \subfigure[]{ 
    \includegraphics[width=0.8\textwidth]{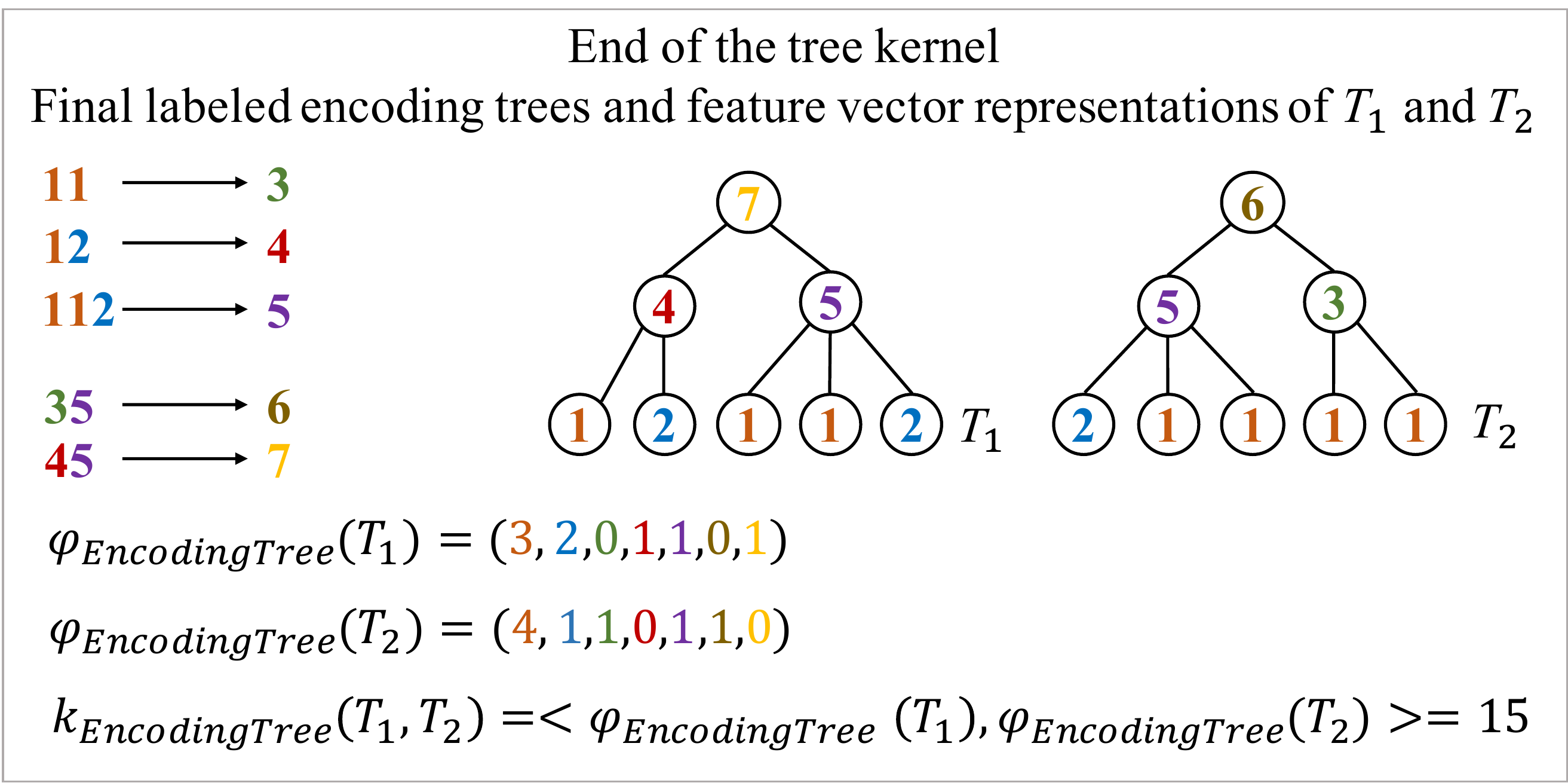} 
    \label{fig:treeKerenl_5} 
  }
  \caption{Illustration of computing the tree kernel on two encoding trees with height=2. Here, $\{1, 2, \dots, 7\}\in \Sigma$ are considered letters.} 
  \label{fig:treeKernel}
\end{figure}

\textbf{\textit{Hierarchical Reporting.}} The key idea of this scheme is to assign labels to non-leaf nodes by aggregating and sorting the labels from their child nodes and then to compress these sorted label sets into new and short labels. Labels from the leaf nodes are iteratively propagated to the root node, which means that the iteration time of this scheme is determined by the height of the encoding tree. See Figure \ref{fig:treeKernel}, a-d, for an illustration of this scheme.

\begin{mydef}
\label{def:tree_kernel}
Let $T_1$ and $T_2$ be any two encoding trees with the same height $h$. There exists a set of letters $\Sigma^i \in \Sigma$, which are node labels appearing at the $i$-th ($i<h$) height of $T_1$ or $T_2$ (i.e., the nodes at the $i$-th height are assigned labels with hierarchical reporting). $\Sigma^0$ is the set of leaf node labels of $T_1$ and $T_2$. Assume that any two $\Sigma^i$ are disjoint, and every $\Sigma^i = \{\evb^i_1,\dots,\evb^i_{\left|\Sigma^i\right|}\}$ is ordered without loss of generality. We define a function $c^i: \{T_1,T_2\}\times\Sigma^i \rightarrow \sB$ such that $c^i(T_1, \evb^i_j)$ counts the number of the letter $\evb^i_j$ in the encoding tree $T_1$.

The tree kernel on the two trees ($T_1$ and $T_2$) with height $h$ after the root nodes are assigned labels is defined as:
\begin{equation}
k_{EncodingTree}(T_1,T_2)=<\varphi_{EncodingTree}(T_1),\varphi_{EncodingTree}(T_2)>,
\end{equation}
where
\begin{equation*}
\varphi_{EncodingTree}(T_1)=(c^0(T_1, \evb^0_1),\dots,c^0(T_1, \evb^0_{\left|\Sigma^0\right|}),\dots,c^h(T_1, \evb^h_1),\dots,c^h(T_1, \evb^h_{\left|\Sigma^h\right|})),
\end{equation*}
and
\begin{equation*}
\varphi_{EncodingTree}(T_2)=(c^0(T_2, \evb^0_1),\dots,c^0(T_2, \evb^0_{\left|\Sigma^0\right|}),\dots,c^h(T_2, \evb^h_1),\dots,c^h(T_2, \evb^h_{\left|\Sigma^h\right|})).
\end{equation*}
\end{mydef}
Following the label counting process in the WL subtree kernel, our tree kernel is also designed to count the number of common labels in two encoding trees. An illustration of this kernel is shown in Figure \ref{fig:treeKernel}.

\begin{mytheo}
\label{theo:treeKernelComplexity}
WL-ET kernel on two encoding trees $T_1$ and $T_2$ with the same height $h$ can be computed in time $O(n)$, which is much simpler than the WL subtree kernel ($O(hm)$) with $h$ iterations on $m$ edges \citep{shervashidze2011weisfeiler} and is the simplest method in graph classification to the best of our knowledge \citep{wu2020comprehensive}.
\end{mytheo}

\begin{myproof}
Given a graph $G$, the biggest encoding tree is a binary encoding tree. Thus, the complexity of WL-ET kernel on the binary encoding tree is the worst case (i.e., $O_{EncodingTree}\leq O_{BinaryEncodingTree}$). The complexity on the binary encoding tree is calculated as:
\begin{align}
\nonumber
O_{BinaryEncodingTree} & = O(|V_T|) \\
\nonumber & = O(|V_T^0| + |V_T^1|, \dots, +|V_T^k|) \\
\nonumber & \leq O(2n) \\
\nonumber & = O(n) 
\end{align}
\end{myproof}

\subsubsection{Computing the tree kernel on many encoding trees}
In addition to the tree kernel designed for measuring the similarity between encoding trees, we aim to propose an algorithm to compute the feature vectors formed on $N$ encoding trees for classification. As shown in Algorithm \ref{code:tree_kernel}, we present the process for one iteration of our tree kernel computation process on $N$ encoding trees.
This algorithm consists of the same 4 steps as those in the WL test, including multiset-label determination, multiset sorting, label compression and relabeling. The core difference is in step 1, where the multiset-label $L^i(v)$ consists of labels from the child nodes of $v$ rather than its neighbors.
Consistent with the WL test, $\Sigma$ is sufficiently large to make $f$ injective.
In a case with $N$ encoding trees, a $\Sigma$ of size $2nN$ suffices.

\begin{algorithm}[!ht]
\caption{One iteration of the tree kernel computation process on $N$ encoding trees}
\label{code:tree_kernel}
\begin{algorithmic}[1]
\STATE Multiset-label determination
\begin{itemize}[leftmargin=2em]
\item For $i=0$, set $L^i(v) = \evb^0(v)$, where $\evb^0(v)$ is the initial leaf node label.
\item For $i>0$, each node $v$ with height $i$ in an encoding tree $T$ is assigned a multiset label $L^i(v) = \{\evb^{i-1}(u)|u\in\gC(v)\}$, where $\gC(v)$ denotes the children of node $v$ in $T$.
\end{itemize}
\STATE Multiset sorting
\begin{itemize}[leftmargin=2em]
\item The elements in $L^i(v)$ are sorted in ascending order and then united into a string $s^i(v)$.
\end{itemize}
\STATE Label compression
\begin{itemize}[leftmargin=2em]
\item An injective function $f:\Sigma^*\rightarrow\Sigma$ is employed to compress each string $s^i(v)$ into a new and short label.
\end{itemize}
\STATE Relabeling
\begin{itemize}[leftmargin=2em]
\item $\evb^i(v):=f(s^i(v))$ is set for all nodes with height $i$ in $T$.
\end{itemize}
\end{algorithmic}
\end{algorithm}

\begin{mytheo}
For $N$ encoding trees with height $h$, WL-ET kernel on all pairs of these encoding trees can be computed in $O(nN^2)$ \footnote{Again, we compute the runtime complexity with the form of binary encoding tree, which is the upper bound of our normal encoding tree.}, which is much simpler than the WL subtree kernel formed on $N$ graphs ($O(Nhm+N^2hn)$) \citep{shervashidze2011weisfeiler} and is the simplest graph classification method.
\end{mytheo}

\begin{myproof}
The runtime complexity of WL-ET kernel with a naive application derived by computing an $N \times N$ kernel matrix is $O(nN^2)$.
\end{myproof}

\subsection{ETL for graph classification}
Based on our tree kernel, we develop a novel deep learning architecture, \textit{Encoding Tree Learning (ETL)}, that generalizes the hierarchical reporting scheme to update the hidden features of non-leaf nodes for graph classification.
ETL uses the tree structure and leaf node features $X_v$ to learn the representation vector of the entire graph $r_T$. ETL follows the proposed hierarchical reporting scheme,
where the representation of a non-leaf node is updated by aggregating the hidden features of its children.
Formally, the $i$-th layer of ETL is
\begin{equation}
\label{eq:etl_aggre}
r_v^i = \text{MLP}^i\left(\sum\nolimits_{u\in\gC(v)}r_u^{(i-1)}\right).
\end{equation}
where $r^i_v$ is the feature vector of node $v$ with height $i$ in the encoding tree. We initialize $r^0_v = X_v$, and $\gC(v)$ is a set of child nodes of $v$.
As shown in Equation \ref{eq:etl_aggre}, we employ summation and multilayer perceptrons (MLPs) to perform hierarchical reporting in ETL. Sum aggregators have been theoretically proven to be injective over multisets and more powerful than mean and max aggregators \citep{xu2019powerful}. MLPs are capable of representing the compositions of functions because of the universal approximation theorem \citep{hornik1989multilayer,hornik1991approximation}.

For graph classification, the root node representation $r^h_v$ can be naively employed as the representation of the entire encoding tree $r_T$.
However, as discussed in \citet{xu2019powerful}, better results could be obtained from features in earlier iterations.
To cover all hierarchical information, we employ hidden features from each height/iteration of the model. This is achieved by an architecture similar to the GIN \citep{xu2019powerful}, where we model the entire encoding tree with layer representations concatenated across all heights/layers of the ETL structure:
\begin{equation}
\label{eq:eml_readout}
r_T = \text{CONCAT}\left(\text{LAYERPOOL}(\{r_v^i | v\in T^i\}) | i=0,1,\dots,h\right),
\end{equation}
where $r_v^i$ is the feature vector of node $v$ with height $i$ in encoding tree $T$ and $h$ is the height of $T$. In ETL, LAYERPOOL in Equation \ref{eq:eml_readout} can be replaced with the summation or averaging of all node vectors within the same iterations as in GIN, which provably generalizes our tree kernel.

\section{Comparisons to related work}
We now discuss why the structural optimization is capable of making graph classification simpler and better.

The simplicity is mainly reflected in three aspects. First, we employ structural optimization to transform the given data sample from a graph into an encoding tree, which is a simpler data structure for representation.
Second, based on the transformed encoding trees, our proposed tree kernel is much simpler than the most popular graph kernel, the WL subtree kernel \citep{shervashidze2011weisfeiler}, for graph classification in terms of runtime complexity. As seen in Theorem \ref{theo:treeKernelComplexity}, our tree kernel for the similarity measurement of a pair of encoding trees can be computed in time $O(n)$, which is naturally less than the $O(hm)$ required for running the WL subtree kernel.
Finally, regarding the use of deep learning models for graph classification, our proposed ETL also achieves a lower complexity than previous GNNs, such as the graph convolutional network (GCN) \citep{kipf2017semi} and GIN \citep{xu2019powerful}. Considering that ETL is generalized from our tree kernel, the runtime complexity of ETL for learning an encoding tree is $O(n)$, while at least $O(hm)$ is required for previous GNNs to learn a graph \citep{wu2020comprehensive}.

The great success of deep networks is attributed to their excellent feature characterization power; put differently, ``they exploit a particular form of compositionality in which features in one layer are combined in many different ways to create more abstract features in the next layer'' \citep{bengio2021deep}.
Regarding graph representation learning, the expressive power of GNNs has also been theoretically proven \citep{xu2019powerful}.
To achieve better graph classification, we optimize the feature characterization ability, including feature extraction and combination, of kernel-based and deep learning-based methods.
In particular, through structural optimization, we transform a graph into a powerful data structure, an encoding tree, which is simple but helpful for feature extraction and combination.
Specifically, we first optimize the feature extraction process.
The encoding tree optimized from a graph has the minimum structural entropy and decodes the key structure underlying the original graph. Following the hierarchical structure of the encoding tree, the features extracted for each tree node are hierarchical, while GNNs are designed to learn representations of the original graph nodes in a flat space.
For feature combination, we optimize the form of compositionality in different iterations/layers. With the hierarchical reporting scheme, the features are combined from bottom to top according to the hierarchical relationships among tree layers, which is obviously different from the form of message passing in GNNs.

\section{Experiments}
We validate the effectiveness of structural optimization by comparing the experimental results of our tree kernel and ETL with those of the most popular kernel-based methods and GNNs on graph classification tasks \footnote{The code of WL-ET kernel and ETL can be found at \url{https://github.com/BUAA-WJR/SO-ET}.}.

\subsection{Datasets}
We conduct graph classification on 5 benchmarks: 3 social network datasets (IMDB-BINARY, IMDB-MULTI, and COLLAB) and 2 bioinformatics datasets (MUTAG and PTC) \citep{xu2019powerful} \footnote{Considering the limitations of encoding trees on disconnected graphs, we utilize additional configurations for the other 4 datasets that contain such graphs. The results are presented in Appendix \ref{app:disconnected_experiment}.}.
There is a difference between the data of bioinformatic datasets and social network datasets; that is, the nodes in bioinformatics graphs have categorical labels that do not exist in social networks.
Thus, the initial node labels for the tree kernel are organized as follows: the node degrees are taken as node labels for social networks; the combination of node degrees and node categorical labels are taken for bioinformatic graphs.
Correspondingly, the initial node features of the ETL inputs are set to one-hot encodings of the node degrees for social networks and a combination of the one-hot encodings of the degrees and categorical labels for bioinformatic graphs.
Table \ref{tab:test_acc} summarizes the characteristics of the 5 employed datasets, and detailed data descriptions are shown in Appendix \ref{sec:data_des}.

\subsection{Configurations}
Following \citet{xu2019powerful}, 10-fold cross-validation is conducted to make a fair comparison, and we present the average accuracies achieved to validate the performance of our methods in graph classification.
Regarding the configuration of our tree kernel, we adopt the $C$-support vector machine ($C$-SVM) \citep{chang2011libsvm} as the classifier and tune the hyperparameter $C$ of the SVM and the height of the encoding tree $\in[2, 3, 4, 5]$. We implement the classification program with an SVM from Scikit-learn \citep{pedregosa2011scikit}, where we set another hyperparameter $\gamma$ as \textit{auto} for IMDB-BINARY and IMDB-MULTI and as \textit{scale} for COLLAB, MUTAG and PTC and set the other hyperparameters as their default values.

For configuration of ETL, the number of ETL iterations is consistent with the heights of the associated encoding trees, which are also $\in [2, 3, 4, 5]$. All MLPs have 2 layers, as in the setting of the GIN \citep{xu2019powerful}. For each layer, batch normalization is applied to prevent overfitting. We utilize the Adam optimizer and set its initial learning rate to 0.01. For a better fit, the learning rate is decayed by  half every 50 epochs. Other tuned hyperparameters for ETL include the number of hidden dimensions $\in \{16, 32, 64\}$; the minibatch size $\in\{32, 128\}$; the dropout ratio $\in\{0, 0.5\}$ after the final output; the number of epochs for each dataset is selected based on the best accuracy within cross-validation results.
We apply the same layer-level pooling approach (LAYERPOOL in Eq. \ref{eq:eml_readout}) for ETL; specifically, sum pooling is conducted on the bioinformatics datasets, and mean pooling is conducted on the social datasets due to better test performance.

\subsection{Baselines}
We compare our tree kernel and ETL model configured above with several state-of-the-art baselines for graph classification: (1) kernel-based methods, i.e., the WL subtree kernel \citep{shervashidze2011weisfeiler} and Anonymous Walk Embeddings (AWE) \citep{ivanov2018anonymous}; (2) state-of-the-art deep learning methods, i.e., Diffusion-Convolutional Neural Network (DCNN) \citep{atwood2016diffusion}, PATCHY-SAN \citep{niepert2016learning}, Deep Graph CNN (DGCNN) \citep{zhang2018end} and GIN \citep{xu2019powerful}. The accuracies of the WL subtree kernel are derived from \citet{xu2019powerful}. For AWE and the deep learning baselines, we utilize the accuracies contained in their original papers.

\renewcommand\arraystretch{1.2}
\begin{table}[!ht]
\centering
\caption{\textbf{Classification accuracies on 5 benchmarks (\%).} The best results are highlighted in boldface. On datasets where WL-ET and ETL are not strictly the highest-scoring models among the baselines, our methods still achieve competitive results; thus, their accuracies are still highlighted in boldface. For the results of the baselines, we highlight those that are significantly higher than those of all other methods.}
\label{tab:test_acc}
\begin{tabular}{lccccc}
\hline
Dataset & IMDB-B & IMDB-M & COLLAB & MUTAG & PTC \\
\# Graphs & 1000 & 1500 & 5000 & 188 & 344 \\
\# Classes & 2 & 3 & 3 & 2 & 2 \\
Avg. \# Nodes & 19.8 & 13.0 & 74.5 & 17.9 & 25.5 \\ \hline
\multicolumn{6}{c}{Kernel-based methods} \\ \hline
WL & 73.8$\pm$3.9 & 50.9$\pm$3.8 & 78.9$\pm$1.9 & 90.4$\pm$5.7 & 59.9$\pm$4.3 \\
AWE & 74.5$\pm$5.9 & 51.5$\pm$3.6 & 73.9$\pm$1.9 & 87.9$\pm$9.8 &  \\
\textbf{WL-ET} & \textbf{74.7$\pm$3.5} & \textbf{52.4$\pm$4.5} & \textbf{81.5$\pm$1.2} & \textbf{89.5$\pm$6.1} & \textbf{63.7$\pm$4.7} \\ \hline
\multicolumn{6}{c}{Deep learning methods} \\ \hline
DCNN & 49.1 & 33.5 & 52.1 & 67.0 & 56.6 \\
PATCHY-SAN & 71.0$\pm$2.2 & 45.2$\pm$2.8 & 72.6$\pm$2.2 & \textbf{92.6$\pm$4.2} & 60.0$\pm$4.8 \\
DGCNN & 70.0 & 47.8 & 73.7 & 85.8 & 58.6 \\
GIN-0 & 75.1$\pm$5.1 & 52.3$\pm$2.8 & 80.2$\pm$1.9 & 89.4$\pm$5.6 & 64.6$\pm$7.0 \\
\textbf{ETL} & \textbf{76.7$\pm$4.5} & \textbf{53.1$\pm$4.5} & \textbf{81.8$\pm$1.2} & \textbf{90.6$\pm$6.8} & \textbf{66.3$\pm$4.3} \\ \hline
\end{tabular}
\end{table}

\section{Results}
The results of validating our tree kernel and ETL model on graph classification tasks are presented in Table \ref{tab:test_acc}. Our methods are shown in boldface.
In the panel of kernel-based methods, we can observe that the accuracies of our tree kernel exceed those of other kernel-based methods on 4 out of 5 benchmarks. For the only failed dataset, MUTAG, our tree kernel still achieves very competitive performance.
Notably, our tree kernel even outperforms the state-of-the-art deep learning method (i.e., GIN-0) on IMDB-MULTI, COLLAB and MUTAG, which implies that superior performance can sometimes be obtained through an optimization method rather than a deep learning method.

In the lower panel containing the deep learning methods, we can observe that the results of ETL are naturally superior to the accuracies of the tree kernel, which further confirms the outperformance of deep learning in terms of feature characterization. In addition, ETL also yields the best performance on 4 out of 5 datasets, while competitive performance can still be observed on the other dataset (i.e., MUTAG). These results indicate that optimization methods can not only coexist with but also further boost deep learning methods. We also compare the volumes of computations of ETL and GIN-0 in Appendix \ref{sec:flops_com}, and the results show that ETL requires only 22\% of the volume of computations used in the GIN on average.

\section{Conclusion and future work}
In this paper, to boost the performance of basic models while simplifying its learning process, we propose structural optimization, which is a structural transformation from original datasets to a simplified new structure that preserves the key features of the input data. Utilizing a recently developed structural entropy minimization algorithm, we improve upon the graph classification performance by simplifying the corresponding kernel method and deep learning method. In particular, our proposed tree kernel and ETL make graph classification simpler and better with optimized encoding trees.
In addition to the excellent graph classification performance, the ETL that derived from structural optimization even possess powerful interpretability with respect to node importance and feature combination paths because of the hierarchical structure of the constructed encoding tree \footnote{An illustration of the underlying interpretability of ETL is shown in Appendix \ref{sec:interpret}.}.
Thus, an interesting direction for future work is to interpret the power of ETL.
Furthermore, despite the superior performance of our proposed methods in graph classification, they are not fit for another important task in graph realm, i.e., node classification. Hence, how structural optimization makes node classification simpler and better may be another underlying direction for future work.

\subsubsection*{Acknowledgments}
This research was supported by National Natural Science Foundation of China under grant 61932002.

\bibliography{my_ref}
\bibliographystyle{so_preprint}

\newpage
\appendix
\section{Experiments on datasets with disconnected graphs}
\label{app:disconnected_experiment}
Considering the limitation of encoding tree on disconnected graphs, we take additional configurations for the other 4 datasets that contain disconnected graphs.

\subsection{Datasets}
There are 4 other well-known graph classification benchmarks with disconnected graphs: 2 bioinformatics datasets (PROTEINS and NCI1) and 2 social network datasets (REDDIT-BINARY and REDDIT-MULTI5K). The input features for these 4 datasets are consistent with the feature handling approach in the main text.
Notably, we take additional configurations for the disconnected graphs contained in these 4 datasets. (1) We transform each connected component from a disconnected graph sample separately into corresponding encoding trees, (2) and then, we combine these separate encoding trees into one tree through naive root node merging.
Table \ref{tab:test_acc_dis} summarizes the data statistics of the adopted benchmarks, and detailed data descriptions are shown in Appendix \ref{sec:data_des}.

\subsection{Models and configurations}
Following the setting in the main text, 10-fold cross-validation is conducted to make a fair comparison, and we present the average accuracies obtained to validate the performance of our methods on graph classification tasks.
Regarding the configuration of our tree kernel, we also tune the hyperparameter $C$ of the SVM and the height of the encoding tree $\in[2, 3, 4, 5]$. We set the other hyperparameter $\gamma$ as \textit{auto} for REDDIT-BINARY and REDDIT-MULTI5K and as \textit{scale} for PROTEINS and NCI1. The configuration of ETL on these 4 datasets is consistent with that in the main text.

\subsection{Results}
We compare our methods with the same baselines and report the results in Table \ref{tab:test_acc_dis}. Our methods only achieve superior performance on one dataset (PROTEINS), where disconnected graphs occupy a 4\% proportion of the data.
For the other three datasets, WL has the best performance among all GNN-based models on NCI1, and GIN-0 obtains the highest accuracies on the REDDIT datasets. One explanation for this phenomenon is that the structural information underlying the disconnected graphs can hardly be captured by the methods that are based on structural optimization.

\begin{table}[!ht]
\centering
\caption{Classification accuracies on datasets with disconnected graphs (\%).}
\label{tab:test_acc_dis}
\begin{tabular}{lcccc}
\hline
Dataset & RDT-B & RDT-M5K & PROTEINS & NCI1 \\
\# Graphs & 2000 & 5000 & 1113 & 4110 \\
\# Disconnected Graphs & 1022 & 3630 & 46 & 580 \\
\# Classes & 2 & 5 & 2 & 2 \\
Avg. \# Nodes & 429.6 & 508.5 & 39.1 & 29.8 \\ \hline
\multicolumn{5}{c}{Kernel-based methods} \\ \hline
WL & 81.0$\pm$3.1 & 52.5$\pm$2.1 & 75.0$\pm$3.1 & \textbf{86.0$\pm$1.8} \\
AWE & 87.9$\pm$2.5 & 54.7$\pm$2.9 &  &  \\
\textbf{WL-ET} & \textbf{86.9$\pm$6.1} & \textbf{53.3$\pm$2.4} & \textbf{76.2$\pm$3.3} & 76.5$\pm$3.3 \\ \hline
\multicolumn{5}{c}{Deep learning methods} \\ \hline
DCNN &  &  & 61.3 & 62.6 \\
PATCHSAN` & 86.3$\pm$1.6 & 49.1$\pm$0.7 & 75.9$\pm$2.8 & 78.6$\pm$1.9 \\
DGCNN &  &  & 75.5 & 74.4 \\
GIN-0 & \textbf{92.4$\pm$2.5} & \textbf{57.5$\pm$1.5} & 76.2$\pm$2.8 & 82.7$\pm$1.7 \\
\textbf{ETL} & 86.8$\pm$1.9 & 51.9$\pm$2.6 & \textbf{76.5$\pm$2.5} & 79.3$\pm$1.4 \\ \hline
\end{tabular}
\end{table}

\section{Details of the datasets}
\label{sec:data_des}
Here, we present detailed descriptions of the 9 benchmarks utilized in this paper.

\textbf{Social network datasets.} IMDB-BINARY and IMDB-MULTI are movie collaboration datasets. Each graph corresponds to an ego network for each actor/actress, where the nodes correspond to actors/actresses and an edge is drawn between two actors/actresses if they appear in the same movie. Each graph is derived from a prespecified genre of movies, and the task is to classify the genre from which each graph is derived. REDDIT-BINARY and REDDIT-MULTI5K are balanced datasets, where each graph corresponds to an online discussion thread and nodes correspond to users. An edge is drawn between two nodes if at least one of them responds to another's comment. The task is to classify each graph into the community or subreddit to which it belongs. COLLAB is a scientific collaboration dataset derived from 3 public collaboration datasets, namely, High Energy Physics, Condensed Matter Physics and Astro Physics. Each graph corresponds to an ego network of a different researcher from each field. The task is to classify each graph into a field to which the corresponding researcher belongs.

\textbf{Bioinformatics datasets.} MUTAG is a dataset containing 188 mutagenic aromatic and heteroaromatic nitro compounds with 7 discrete labels. PROTEINS is a dataset where the nodes are secondary structure elements (SSEs), and there is an edge between two nodes if they are neighbors in the given amino acid sequence or in 3D space. The dataset has 3 discrete labels, representing helixes, sheets or turns. PTC is a dataset containing 344 chemical compounds that reports the carcinogenicity of male and female rats and has 19 discrete labels. NCI1 is a dataset made publicly available by the National Cancer Institute (NCI) and is a subset of balanced datasets containing chemical compounds screened for their ability to suppress or inhibit the growth of a panel of human tumor cell lines; this dataset possesses 37 discrete labels.

\begin{figure}[!hb]
  \centering
  \includegraphics[width=1.\textwidth]{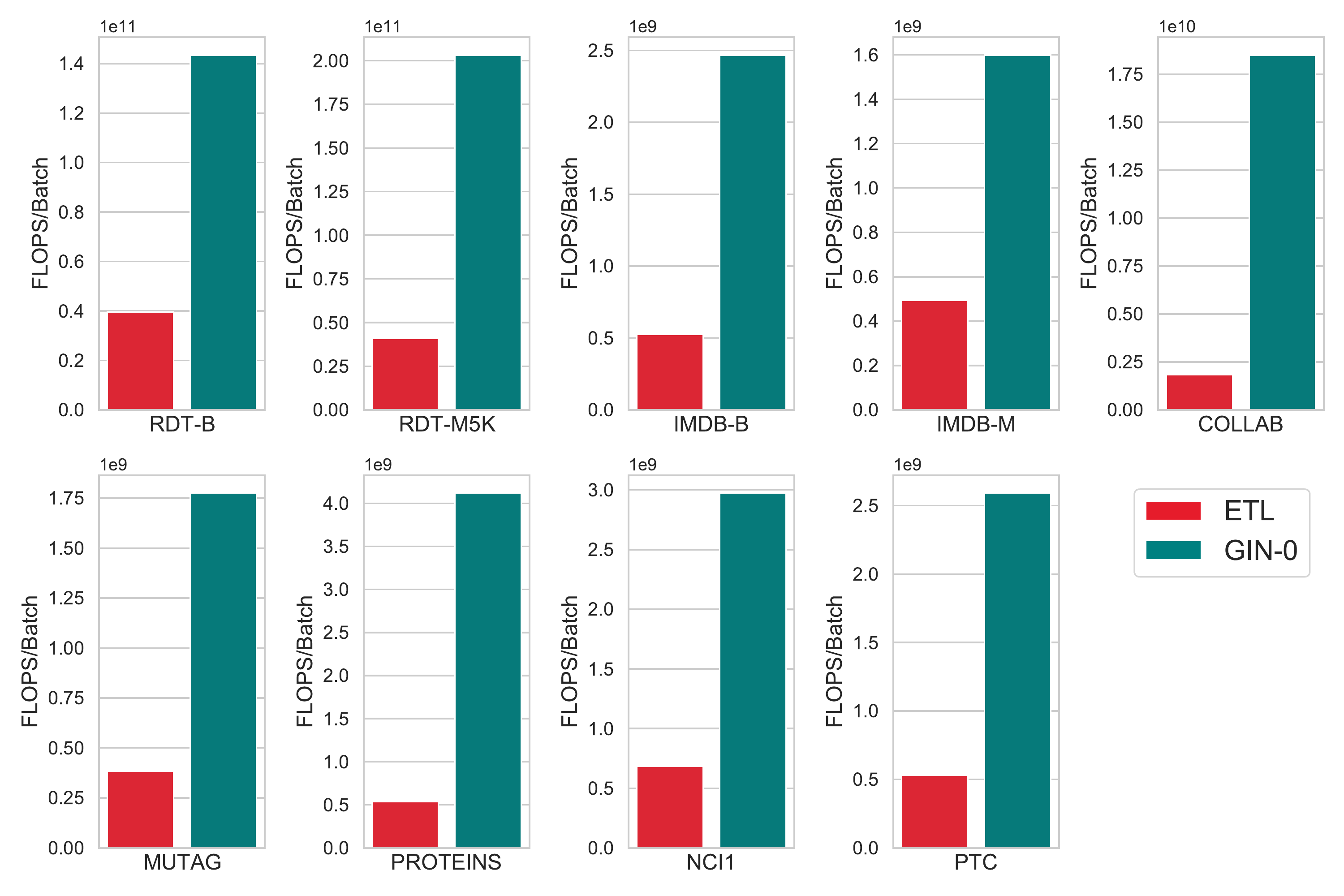}   
  \caption{Comparison regarding the required volume of computations.} 
  \label{fig:log_flops}
\end{figure}

\section{Analysis of the efficiency of ETL}
\label{sec:flops_com}
In addition to the lower runtime complexity of our ETL approach than that of state-of-the-art GNNs ($O(n) < O(hm)$), we also compare the volumes of computations, i.e., the numbers of floating-point operations per second (FLOPS), required by ETL and GIN-0 under the same parameter settings on all datasets. Specifically, we fix the number of iterations to 4 (the 5 GNN layers of GIN-0 include the input layer), the number of hidden units to 32, the batch size to 128 and the final dropout ratio to 0.
The results are shown in Figure \ref{fig:log_flops}. One can see that the volume of computations required by our ETL method is consistently smaller than that of GIN-0. More concretely, ETL needs only 22\% of the volume of computations needed by the GIN on average.

\section{The running time of structural optimization}
\label{sec:time_com}
\begin{figure}[!hb]
  \centering
  \includegraphics[width=1\textwidth]{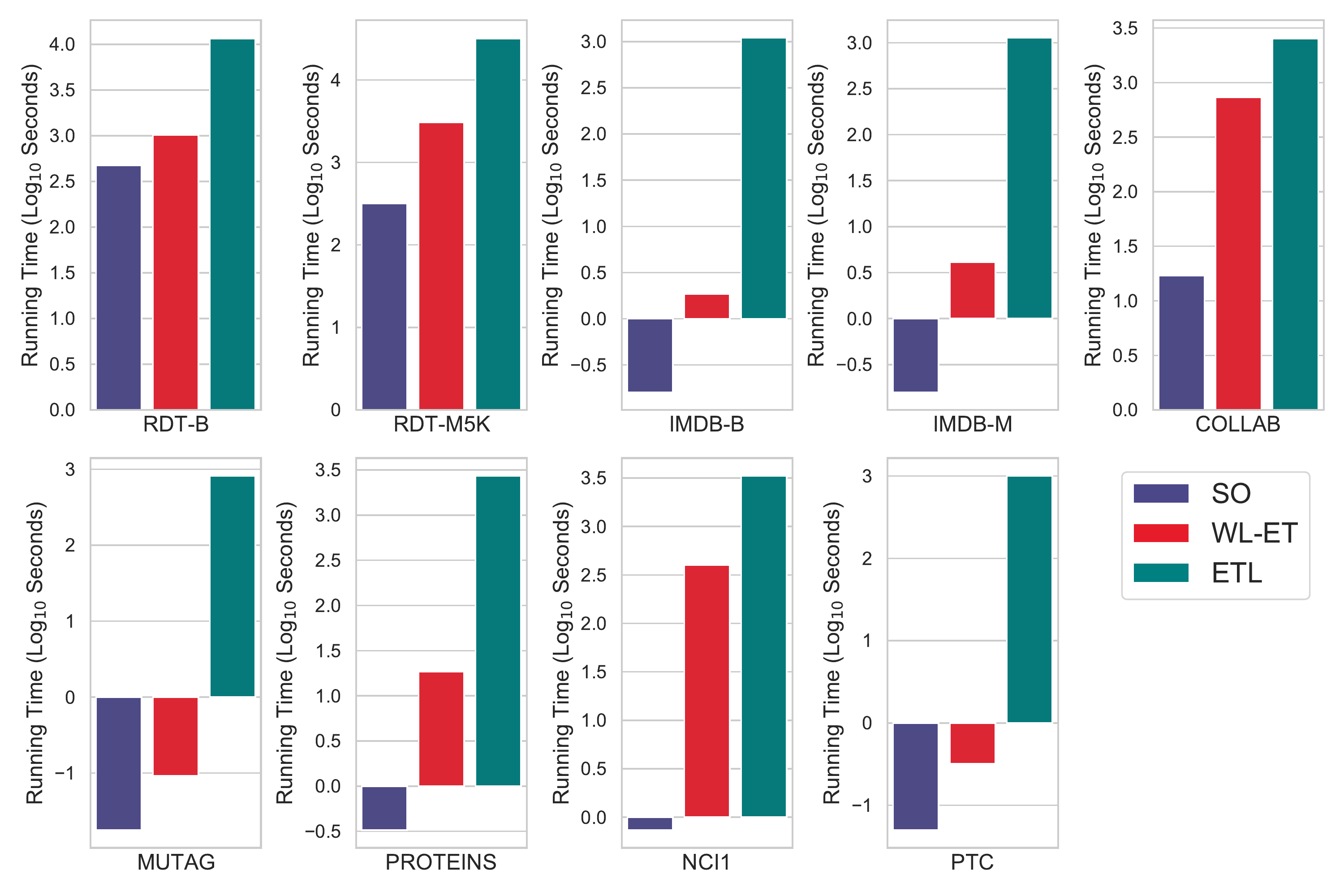}   
  \caption{\textbf{Running time comparison}. Structural optimization and WL-ET are conducted on a machine with an AMD Ryzen 3900x and 64 GB of RAM. ETL is trained with a Tesla V100 GPU. The Y axis is the log with a base of 10 seconds because of the very large distinction in scale.}
  \label{fig:log_time}
\end{figure}

The total time required for generating a classifier includes the time of structural optimization and the training time of ETL, in which the structural optimization only needs to run once, while ETL needs hundreds of epochs of training before its testing even with fixed hyperparameters.
Thus, we compare the real running times of structural optimization (SO), WL-ET \footnote{The running time of WL-ET is the time required for performing 10-fold cross-validation with $C$-SVM.} and ETL \footnote{We calculate the actual running time of ETL with fixed hyperparameters under 300 epochs training.} on all datasets with the fixed hyperparameters described in Section \ref{sec:flops_com}. The results are shown in Figure \ref{fig:log_time}, and we can see that the time required for structural optimization is much less than the time needed by WL-ET and ETL and only accounts for 0.002\% to 1\% of the time required by ETL.

\section{Underlying interpretability advantage}
\label{sec:interpret}
Compared with GNNs, our ETL is more interpretable regarding its node importance levels and feature combination paths. Based on the optimized encoding trees, the features propagating in ETL follow a single direction rather than the complicated data loops found in graphs, which makes it hard for GNNs to interpret the entire information propagation path.

\begin{figure}[!hb]
  \centering
  \includegraphics[width=0.8\textwidth]{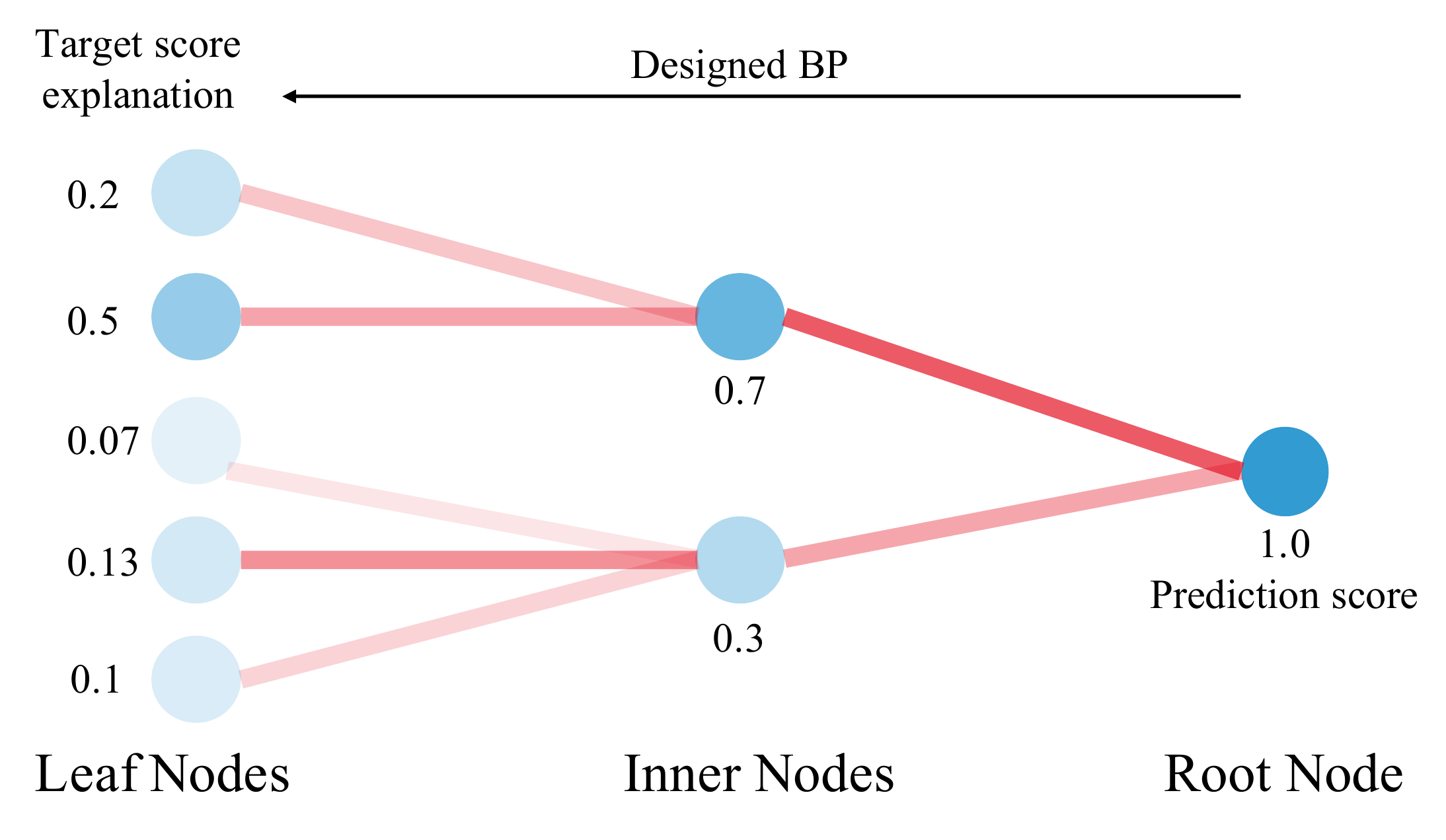}   
  \caption{An example of the interpretability of ETL with a decomposition method.} 
  \label{fig:etl_interpret}
\end{figure}

Following the decomposition methods used in GNNs, we can not only measure the importance levels of nodes with different heights but also identify the most important path for feature combination by combining the importance scores of the nodes in the path. A toy example to demonstrate the advantage of ETL in terms of interpretability can be found in Figure \ref{fig:etl_interpret}.
As in the decomposition methods of GNNs. The intuition of this idea is to build score decomposition rules to distribute the prediction scores from the output space to the input space.
Starting from the output layer, the model's prediction is treated as the initial target score. Then, the score is decomposed and distributed to the neurons in the previous layer following the decomposition rules. By repeating such procedures until covering the input space, the importance scores for node features can be obtained, which can also be combined to represent path importance.

\end{document}